\documentclass[letterpaper, 10 pt, conference]{ieeeconf}  % Comment this line out if you need a4paper

\IEEEoverridecommandlockouts                              % This command is only needed if 
\makeatletter
\let\NAT@parse\undefined
\makeatother

\usepackage{amsmath} % assumes amsmath package installed
\usepackage{amssymb}  % assumes amsmath package installed
\usepackage{tikz}
\usepackage{bm}
\usepackage{xcolor}
\usepackage{graphicx}
\usepackage[numbers]{natbib}
\usepackage[bookmarks=true]{hyperref}
\usepackage{cleveref}
\usepackage{multirow}
\usepackage{makecell}
\usepackage{booktabs}
\usepackage{todonotes}
\usepackage{chngcntr}
\usepackage{dsfont}
\crefalias{section}{appendix}

\newtheorem{definition}{Definition}
\newcommand{\parents}{\mathrm{Pa}} 

\usepackage{xspace}
\makeatletter
\DeclareRobustCommand\onedot{\futurelet\@let@token\@onedot}
\def\@onedot{\ifx\@let@token.\else.\null\fi\xspace}
\makeatother

\newcommand{\ie}{i.e\onedot}

\newcommand{\algo}[1]{\textsc{#1}}
\newcommand{\method}{\algo{CAIMAN}\xspace}
\def\code#1{\texttt{#1}}

\definecolor{ourblue}{rgb}{0.368,0.507,0.71}
\definecolor{ourorange}{rgb}{0.881,0.611,0.142}
\definecolor{ourgreen}{rgb}{0.56,0.692,0.195}
\definecolor{ourred}{rgb}{0.923,0.386,0.209}
\definecolor{ourviolet}{rgb}{0.7, 0.471, 0.701}
\definecolor{ourbrown}{rgb}{0.772,0.432,0.102}
\definecolor{ourlightblue}{rgb}{0.364,0.619,0.782}
\definecolor{ourdarkolive}{rgb}{0.572,0.586,0.}
\definecolor{ourdarkred}{rgb}{0.67, 0.22, 0.07}
\definecolor{ourdarkorange}{rgb}{0.71, 0.49, 0.1}
\definecolor{ourdarkblue}{rgb}{0.27, 0.4, 0.58}
\definecolor{ourdarkgreen}{rgb}{0.41, 0.51, 0.15}

\colorlet{caiman}{ourorange}
\colorlet{heuristics}{ourblue}
\colorlet{cailearn}{ourgreen}
\colorlet{caiprior}{ourred}
\colorlet{transfer_scratch}{ourdarkgreen}
\colorlet{transfer_dyn}{ourdarkred}

\usepackage{subcaption}
\usepackage{tablefootnote}
\usepackage{pdfpages}
\usepackage[linesnumbered,ruled,vlined]{algorithm2e}
\Crefformat{figure}{#2Fig.~#1#3}
\Crefformat{equation}{#2Eq.~#1#3}

\crefname{algocf}{alg.}{algs.}
\Crefname{algocf}{Algorithm}{Algorithms}
\newcommand{\norm}[1]{\left\lVert#1\right\rVert}

\usepackage{wrapfig}

\usepackage{siunitx}
\title{\LARGE \bf
\method: Causal Action Influence Detection for Sample-Efficient Loco-Manipulation
}
\author{Yuanchen~Yuan$^{1}$, 
        Jin~Cheng$^{2}$, 
        Núria~Armengol~Urpí$^{2}$, 
        Stelian~Coros$^{2}$% <-this % stops a space
\thanks{$^{1}$Department of Mechanical and Process Engineering, 
        ETH Zurich, Switzerland
        {\tt\small  yuayuan@ethz.ch}}%
\thanks{$^{2}$Department of Computer Science, 
        ETH Zurich, Switzerland
        {\tt\small \{first.last\}@ethz.ch}}%
}

\begin{document}

\maketitle
% \thispagestyle{empty}
% \pagestyle{empty}
%%%%%%%%%%%%%%%%%%%%%%%%%%%%%%%%%%%%%%%%%%%%%%%%%%%%%%%%%%%%%%%%%%%%%%%%%%%%%%%%
\begin{abstract}

Enabling legged robots to perform non-prehensile loco-manipulation is crucial for enhancing their versatility.
However, learning behaviors such as whole-body object pushing often necessitates sophisticated planning strategies or extensive task-specific reward shaping.
In this work, we present \method, a practical reinforcement learning framework that encourages the agent to gain \textit{control} over other entities in the environment. \method leverages causal action influence as an intrinsic motivation objective, 
allowing legged robots to efficiently acquire object pushing skills even under sparse task rewards.
We employ a hierarchical control strategy, combining a low-level locomotion module with a high-level policy that generates task-relevant velocity commands and is trained to maximize the intrinsic reward. 
To estimate causal action influence, we learn the dynamics of the environment by integrating a kinematic prior with data collected during training.
We empirically demonstrate \method's superior sample efficiency and adaptability to diverse scenarios in simulation, as well as its successful transfer to real-world systems without further fine-tuning.
A video demo is available at \href{https://www.youtube.com/watch?v=dNyvT04Cqaw}{https://www.youtube.com/watch?v=dNyvT04Cqaw}.

\end{abstract}

%%%%%%%%%%%%%%%%%%%%%%%%%%%%%%%%%%%%%%%%%%%%%%%%%%%%%%%%%%%%%%%%%%%%%%%%%%%%%%%%
\section{INTRODUCTION}

Modern day legged robots showcase impressive versatility, from traversing challenging terrains~\cite{miki2022learning, choi2023learning} to executing agile maneuvers such as backflips and parkour~\cite{katz2019mini, cheng2024extreme}.
Yet as expectations for autonomy grow, enabling these systems to physically interact with their environments remains a central research problem~\cite{ha2024learning, tang2024deep, gu2025humanoid}.
A common strategy to enhance manipulation capabilities is to equip legged robots with external manipulators for prehensile tasks~\cite{bellicoso2019alma, zimmermann2021go, sleiman2021unified},
but such methods are inherently constrained by object size and payload.
Harnessing whole-body motion for non-prehensile manipulation offers a promising alternative, but remains a non-trivial challenge.

Traditional approaches for whole-body manipulation explicitly model robot-object interactions, often relying on complex planning and optimization for coordinating motion~\cite{murooka2015whole}.
These methods require accurate models of both the robot and the environment, which restricts their scalability for high-dimensional systems with complex, stochastic dynamics and limits contact points to predefined regions~\cite{farnioli2016toward, rigo2023contact}.
Learning-based methods offer a more scalable alternative, improving computational efficiency and reducing dependence on precise object estimation~\cite{shi2021circus, cheng2023legs, arm2024pedipulate}.
However, these methods frequently rely on tedious reward shaping or learning curriculums to guide exploration and foster meaningful behavior.
Furthermore, the large exploration space of loco-manipulation tasks often necessitates special treatment, such as learning from behavioral priors ~\cite{ha2024umi, urpi2023efficient} or task-agnostic explorative rewards~\cite{schwarke2023curiosity, zhang2024wococo}, to encourage meaningful engagement with the environment.
% Prior works have also leveraged a hierarchical structure to decouple locomotion and high-level planning~\cite{jeon2023learning}. Training the high-level policy, however, still requires elaborate reward design and shaping. 

\begin{figure} %[t]
    \vspace{0.2cm}
    \centering
    \includegraphics[width=\linewidth]{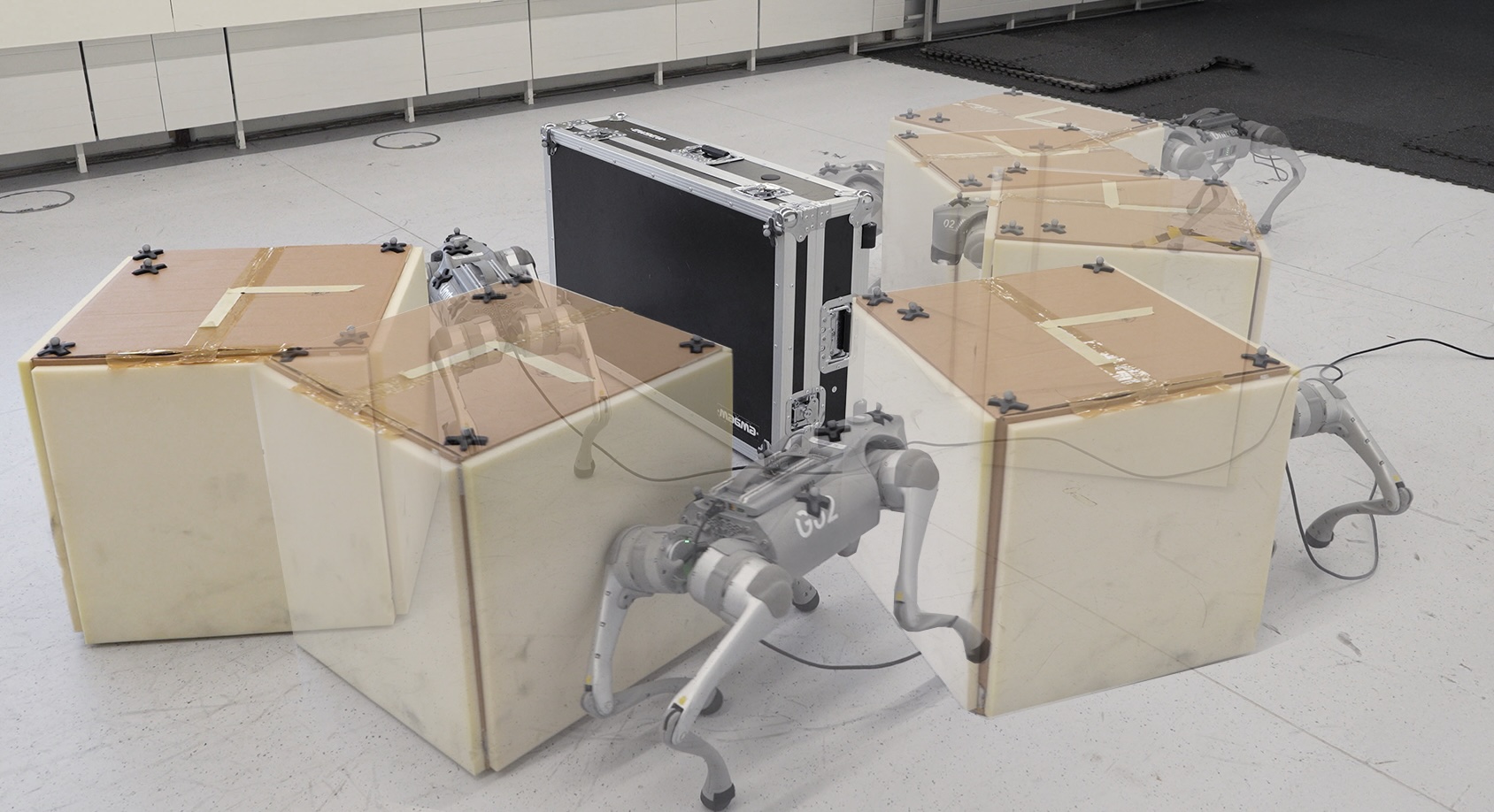}
    \caption{Robot loco-manipulation in the real world enabled by \method. The quadruped maneuvers a box around an obstacle to reach a target position.} 
    \label{fig:teaser}
    \vspace{-0.6cm}
\end{figure}

To alleviate these challenges, we introduce \method, a framework for training non-prehensile object pushing skills in legged robots. 
\method employs a hierarchical control structure that decouples high-level planning from low-level locomotion.
We train robust locomotion using an existing pipeline, and we learn the high-level policy with a simple yet effective reward structure consisting of only three terms: a sparse task reward, an action regularizer, and an intrinsically motivated explorative reward.
The sparse task reward provides a general signal for various pushing tasks, where the robot is only rewarded for completing the task successfully.
The exploratory reward incentivizes the robot to explore and gain control over the environment via Causal Action Influence (CAI)~\cite{seitzer2021causal}, which is a measure quantifying how much influence an agent has on the states of other entities and is computed based on environment dynamics.
% The high-level policy is incentivized to explore and gain control over the environment via an exploratory reward based on Causal Action Influence (CAI)~\cite{seitzer2021causal}, which quantifies how much influence the agent has on the states of other entities and is computed based on environment dynamics.
Instead of learning the dynamics from scratch, \method combines a simple predefined kinematic prior with learned residual dynamics that compensate for physical interactions beyond the prior, allowing accurate dynamics to be learned efficiently.
We show that in a sparse task reward setting, \method achieves strong learning performance and superior sample efficiency compared to other baselines, including within complex, obstacle-laden scenarios.
We also demonstrate that the learned residual accurately captures complex robot-object interactions that are not modeled by the kinematic prior.
Finally, we successfully transfer the learned policy to a real quadruped robot, enabling it to perform whole-body object pushing in real-world settings.

To summarize, our contribution is three-fold: 1) a general hierarchical framework for learning whole-body object pushing with legged robots in various scenarios, including navigation through obstacles; 2) an intrinsically motivated reward based on causal influence calculated from a combination of a kinematics prior and learned residual dynamics; 3) successful hardware validation on a real-world quadruped robot. 

\section{RELATED WORK}
\label{sec:related}
\subsection{Loco-manipulation on Legged Systems}
The integration of locomotion and manipulation has gained significant attention as a promising and application-oriented research field for legged robots.
Model-based methods that rely on accurate representations of both robot and object to optimize trajectories~\cite{zimmermann2021go, murooka2015whole, rigo2023contact} have shown success in tasks such as box carrying~\cite{bellicoso2019alma, sleiman2021unified}, but often require precise state estimation~\cite{lin2024locoman, schakkal2024dynamic} and struggle with scalability due to the complexity of contact modes~\cite{cheng2023enhancing}.
Reinforcement learning (RL) is a viable alternative that avoids explicit modeling and has been successfully applied to diverse tasks including soccer dribbling~\cite{ji2023dribblebot, hu2024dexdribbler}, button pushing~\cite{cheng2023legs, he2024learning}, and door opening~\cite{arm2024pedipulate, schwarke2023curiosity}. However, achieving effective exploration remains difficult given the large search space of robot-object interactions~\cite{shi2021circus, fu2023deep}. To address exploration challenges, researchers have proposed using behavior prior~\cite{sleiman2024guided, liu2024opt2skill} or task-agnostic explorative rewards~\cite{schwarke2023curiosity, zhang2024wococo} as mechanisms to guide learning.
In addition, hierarchical frameworks~\cite{rigo2024hierarchical, wang2024hypermotion} decompose tasks into high-level planning and low-level control, enabling effective behaviors across various scenarios~\cite{kumar2023cascaded}.

In accordance with previous works, we also implement a hierarchical control framework for whole-body object pushing, decoupling high-level velocity planning from low-level locomotion, similarly to Jeon et al.~\cite{jeon2023learning}. 
We aim to utilize a simple reward structure, as opposed to one that requires sophisticated design effort as seen in~\citep{jeon2023learning}, to achieve sample-efficient high-level policy training.

\subsection{Intrinsically Motivated Reinforcement Learning}
Intrinsic motivation (IM) \citep{rm2000intrinsic} plays a vital role in reinforcement learning, especially when extrinsic rewards are sparse or difficult to design. Core IM mechanisms include curiosity~\citep{schmidhuber1991possibility}, learning progress~\citep{schmidhuber2010formal}, and empowerment~\citep{klyubin2005empowerment}, each promoting exploration and skill acquisition.
Curiosity, often measured via prediction errors in learned world models~\citep{pathak2017curiosity, burda2019exploration}, rewards agents for encountering novel states, and has been previously applied for learning loco-manipulation skills~\citep{schwarke2023curiosity, zhang2024wococo}.
Learning progress encourages agents to focus on regions in the state space with rapid improvement, supporting curriculum learning and adaptive exploration~\citep{blaes2019control, colas2019curious}.

Our work builds on empowerment, an information-theoretic quantity defined as the channel capacity between the agent's actions and its sensory observations~\citep{klyubin2005empowerment, mohamed2015variational}. Recent work has connected IM to causal reasoning, aiming to improve sample efficiency and interpretability~\citep{buesing2018woulda, sontakke2021causal}.
To encourage effective exploration, we employ causal action influence (CAI)~\citep{seitzer2021causal}, a measure of an agent's ability to influence its environment and a conceptual lower bound on empowerment. 
\vspace{-0.1cm}

\section{PRELIMINARIES}

\begin{figure}[!t]
    \centering
    \begin{subfigure}[b]{0.9\linewidth}
       \includegraphics[width=1\linewidth]{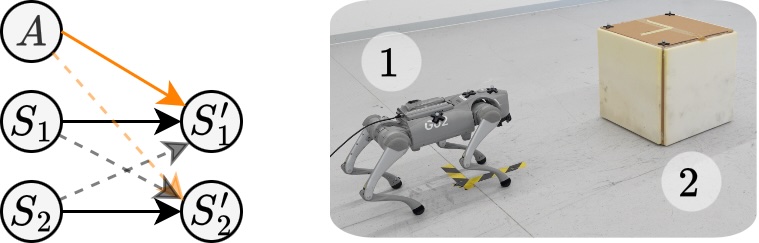}
       \caption{No influence of $A$ on $S'_2$}
        \vspace{0.2cm}
    \end{subfigure}
    \begin{subfigure}[b]{0.9\linewidth}
       \includegraphics[width=1\linewidth]{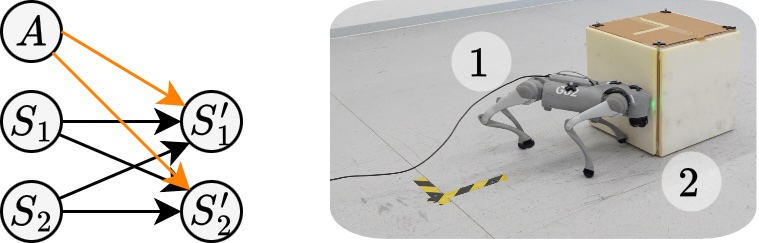}
       \caption{Influence of $A$ on $S'_1$ and $S'_2$}
    \end{subfigure}
    
    \caption{Illustration of the LCM (\textit{left}) for two different environment situations $S=s$ (\textit{right}) in the loco-manipulation task. 
    The LCM captures the transition from $S,A$ to $S'$, factorized into state components. 
    The global SCM is fully connected (dashed and continuous lines), while the LCM $\mathcal{G}_{S=s}$ (continuous lines) is sparser.
    We are interested in detecting the presence of continuous orange arrows in the LCM, \ie the influence of the action $A$ on next states $S'$.} 
    \label{fig:scm}
    \vspace{-0.6cm}
\end{figure}

We model decision-making in a dynamic environment as a Markov Decision Process (MDP)~\citep{sutton2018reinforcement}, defined by the tuple $\langle \mathcal{S}, \mathcal{A}, P, R, \gamma \rangle$, where $\mathcal{S}$ is the state space, $\mathcal{A}$ the action space, $P$ the transition kernel, $R$ the reward function, and $\gamma$ the discount factor.
Following the principle of independent causal mechanisms~\citep{peters2017elements}, we assume that the world consists of interacting but independent entities. This induces a state space factorization $\mathcal{S} = \mathcal{S}_1 \times \cdots \times \mathcal{S}_N$ for $N$ entities, where each factor $\mathcal{S}_i$ represents the state of entity $i$.
An MDP coupled with a policy $\pi: \mathcal{S} \mapsto \mathcal{A}$ induces a \emph{Structural Causal Model} (SCM)~\citep{pearl2009causality} describing the resulting trajectory distribution. 

\begin{figure*}[ht!]
    % \vspace{0.2cm}
    \centering
    \includegraphics[width=\linewidth]{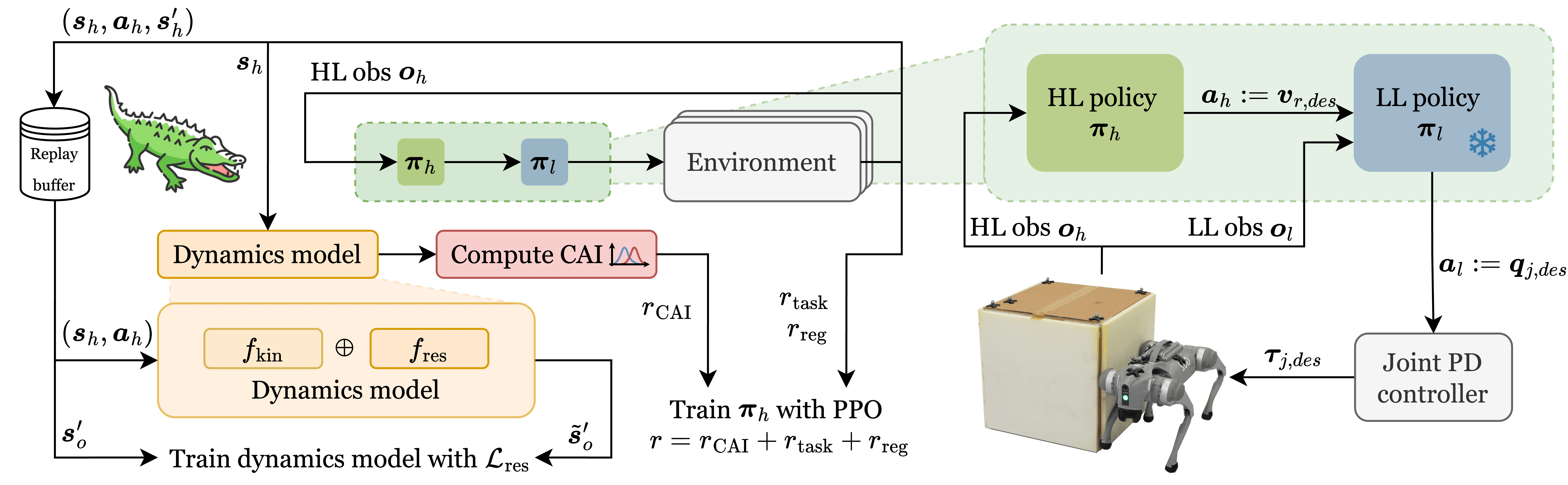}
    \caption{\method framework: The high-level (HL) policy generates desired base velocity commands, which are translated into joint commands by a low-level (LL) policy. 
    We utilize a simple kinematic prior and learned residual dynamics to model the robot-object interaction in the environment while providing a CAI-based explorative bonus along with the sparse task reward. 
    }
    \label{fig:framework}
    \vspace{-0.6cm}
\end{figure*}

\begin{definition}[Structural Causal Model~\citep{pearl2009causality}] 
An SCM is a tuple $(\mathcal{U}, \mathcal{V}, F, P^u)$, where $\mathcal{U}$ is a  set of exogenous variables (e.g., latent randomness) sampled from $P^u$, $\mathcal{V}$ is a set of observed variables (e.g., states, actions, rewards), and $F$ is the set of structural functions capturing the causal relations, such that functions $f_V: \parents(V) \times \mathcal{U} \rightarrow V$, with $\parents(V) \subset \mathcal{V}$ denoting the set of parents of $V$, determine the value of endogenous variables $V$ for each $V \in \mathcal{V}$.
\end{definition}
SCMs are typically visualized as directed acyclic graphs, where nodes represent variables and edges indicate causal relations, as shown in~\Cref{fig:scm}.
The SCM we consider captures one-step transitions with variables $\mathcal{V} = {S_1, \ldots, S_N, A, S_1', \ldots, S_N'}$.
Due to the Markov property and flow of time, causal dependencies exist only from $(S, A)$ to $S'$. 
The global SCM is generally fully connected, encoding all \emph{possible}, however unlikely, interactions between entities (i.e., $S_i / A \rightarrow S'_j$ for many $i,j$) (see ~\Cref{fig:scm}). However, most of these dependencies are inactive given a specific state configuration. For example, the robot can only influence an object if it is within a certain proximity to it. 
%but an edge exists in the global SCM regardless of the robot's location. 
% While most entity pairs may theoretically interact (i.e., $S_i / A \rightarrow S'_j$ for most $i,j$), interactions often become sparse when we observe a specific state configuration.
To capture such context-specific structure, we use the \emph{Local Causal Model} (LCM)~\citep{pitis2020counterfactual}.
\begin{definition}[Local Causal Model~\citep{pitis2020counterfactual}]\label{def:lcm}
Given an SCM $(\mathcal{U}, \mathcal{V}, F, P^u)$ and an observation $V = v, V \subset \mathcal{V}$, the local SCM is the SCM with $F_{V=v}$ and graph $\mathcal{G}_{do(V=v)}$ obtained by pruning edges from $\mathcal{G}_{do(V=v)}$ until it is causally minimal. 

Intuitively, the LCM is sparser than the global SCM, as it retains only the edges corresponding to entity influences that are active in the current context. This ensures that the graph reflects the agent’s \emph{local influences}, rather than all potential ones.
\end{definition}

In this work, we build on the insight that agents can effectively learn loco-manipulation tasks when encouraged to gain \textit{control} over their environment.
We do so by driving the agent toward states where it can influence other entities, achieved through a principled explicit measure of local influence, 
the Causal Action Influence (CAI)~\cite{seitzer2021causal, urpicausal} measure.
CAI is a state-dependent measure of control that assesses whether an agent's actions can affect the states of other entities, 
corresponding graphically to the presence of an edge $A \rightarrow S'_j$ in the LCM $\mathcal{G}_{do(S=s)}$.
Formally, CAI is defined via point-wise conditional mutual information $I(S_j'; A \mid S = s)$, which equals zero when $S_j'$ is independent of $A$ given $S = s$, \ie $S_j' \perp\!\!\!\perp A \mid S = s$. 
At state $S=s$, CAI is given by
\begin{align}
\begin{split}
    C^j(s) &:= I(S'_j; A \mid S=s)  \\ &= \mathbb E_{a\sim \pi} \big[ D_{KL} \big( P_{S'_j\mid S=s,A=a}   \;\big|\big| \; P_{S'_j\mid S = s} \big) \big].
    \label{eq:pointwise_cmi}
\end{split}
\end{align}
\vspace{0.1cm}

\section{METHOD}
The hierarchical control and training framework adopted by \method is presented in~\Cref{fig:framework}. 
It consists of a high-level (HL) policy that generates desired base velocity commands and a low-level policy that translates velocity commands into joint-level actions.  
In this work, we introduce a novel, sample-efficient approach for learning the high-level policy, while building upon an existing pipeline~\citep{rudin2022learning} to train robust low-level policies.% that translate velocity commands into joint-level actions.

\subsection{Low-level Locomotion Policy}
The low-level locomotion policy $\boldsymbol{\pi}_l$ is trained to generate target joint positions $\boldsymbol{q}_{j,des} \in \mathbb{R}^{12}$ that track a desired base velocity command $\boldsymbol{v}_{r,des} \in \mathbb{R}^{3}$.
This command, $\boldsymbol{v}_{r,des} = (v^x_{r,des}, v^y_{r,des}, \omega^z_{r,des})$, specifies linear velocities in the longitudinal ($x$) and lateral ($y$) directions and a yaw rate, all in the robot frame.
The desired joint positions $\boldsymbol{q}_{j,des}$ are tracked using joint-level proportional-derivative (PD) controllers to produce the corresponding joint torques $\boldsymbol{\tau}_{j,des} \in \mathbb{R}^{12}$.
The policy is trained with velocity commands sampled uniformly from a predefined range.
To ensure robust sim-to-real transfer and hardware performance during contact-rich tasks like pushing, we apply domain randomization and external perturbations following~\citep{rudin2022learning}.
The complete low-level observation space $\boldsymbol{o}_l$ is shown in~\Cref{tab:obs}.

\begin{table}[t!]
    \centering
    \renewcommand{\arraystretch}{1.1} % Default value: 1
    \begin{tabular}{|c|c|}
        \hline
        \multicolumn{2}{|c|}{\textbf{Low-level observation} $\boldsymbol{o}_l$} \\ \hline
        $_{\mathcal{B}}\boldsymbol{v}_r\in\mathbb{R}^3$ & Robot linear velocity in base frame $\mathcal{B}$ \\ \hline
        $_{\mathcal{B}}\boldsymbol{\omega}_r\in\mathbb{R}^3$   & Robot angular velocity in base frame $\mathcal{B}$  \\ \hline
        $\boldsymbol{q}_j\in\mathbb{R}^{12}$   & Joint positions   \\ \hline
        $\Dot{\boldsymbol{q}}_j\in\mathbb{R}^{12}$   & Joint velocities   \\ \hline
        $_{\mathcal{B}}\boldsymbol{g}\in\mathbb{R}^3$   & Projected gravity in base frame $\mathcal{B}$ \\ \hline
        %\boldsymbol{v}_{r,des}=(v^x_{r,des}, v^y_{r,des}, \omega^z_{r,des})\in\mathbb{R}^3$   & Desired velocity command \\ \hline
        $\boldsymbol{v}_{r,des}\in\mathbb{R}^3$   & Desired velocity command \\ \hline
        $\boldsymbol{a}_{l,prev}\in\mathbb{R}^{12}$& Previous action  \\ \hline\hline
        
        \multicolumn{2}{|c|}{\textbf{High-level observation} $\boldsymbol{o}_h$ (in world frame $\mathcal{W})$} \\ \hline
        $\boldsymbol{v}_r\in\mathbb{R}^3$   & Robot linear velocity \\ \hline
        $\boldsymbol{\omega}_r\in\mathbb{R}^3$   & Robot angular velocity \\ \hline
        $\boldsymbol{\xi}_r = (x_r, y_r, \psi_r)\in\mathbb{R}^3$   & Robot pose\\ \hline
        $\boldsymbol{\xi}_o = (x_o, y_o, \psi_o)\in\mathbb{R}^3$   & Object pose\\ \hline
        $\boldsymbol{p}_{t} = (x_t, y_t)\in\mathbb{R}^2$   & Target position\\ \hline
        $\boldsymbol{a}_{h, prev}\in\mathbb{R}^3$   & Previous action\\ \hline
        \multicolumn{2}{|l|}{\textit{additional:}} \\ \hline
        $(x_w, y_w, \psi_w)\in\mathbb{R}^3$   & Wall pose (for each wall)\\ \hline\hline
        
        \multicolumn{2}{|c|}{\textbf{High-level state for CAI} $\boldsymbol{s}_h$ (in world frame $\mathcal{W}$)} \\ \hline
        $\boldsymbol{\xi}_r = (x_r, y_r, \psi_r)\in\mathbb{R}^3$   & Robot pose\\ \hline
        $\boldsymbol{\xi}_o = (x_o, y_o, \psi_o)\in\mathbb{R}^3$   & Object pose\\ \hline
        $\boldsymbol{v}_r = (v_r^x, v_r^y, \omega_r^z)\in\mathbb{R}^3$   & Robot velocity \\ \hline
        $\boldsymbol{v}_o = (v_o^x, v_o^y, v_o^z)\in\mathbb{R}^3$   & Object velocity \\ \hline
    \end{tabular}
    \caption{Detailed observation space for each module.}
    \label{tab:obs}
    \vspace{-0.6cm}
\end{table}

\subsection{High-level Planning Policy}
The high-level policy $\boldsymbol{\pi}_h$ is trained to generate desired robot velocity commands $\boldsymbol{a}_h := \boldsymbol{v}_{r,des}$ that achieve successful task completion in object pushing. 
Its observation $\boldsymbol{o}_h$ includes the robot's linear and angular velocities, the poses of the robot and object, the target object position $\boldsymbol{p}_{t} = (x_p, y_p) \in \mathbb{R}^2$, and the previous action---all expressed in the world frame.
In scenarios with obstacles (e.g., walls), $\boldsymbol{o}_h$ also includes their poses in the world frame.
The complete high-level observation space $\boldsymbol{o}_h$ is shown in~\Cref{tab:obs}.

We use Proximal Policy Optimization (PPO)~\cite{schulman2017proximal} as the base RL algorithm, with a simple reward function composed of three terms:
\begin{equation}
r = w_1\mathds{1}_{\norm{\boldsymbol{p}_{o} - \boldsymbol{p}_{t}}_2 < \epsilon} + w_{2} r_{\text{CAI}} + w_3\norm{\boldsymbol{a}_h - \boldsymbol{a}_{h, prev}}_2^2,
\label{eq: rewards}
\end{equation}
where $\boldsymbol{p}_o$, $\boldsymbol{p}_t$ are the current and target object positions, $\epsilon$ is a success threshold, and $\boldsymbol{a}_h$, $\boldsymbol{a}_{h,prev}$ are the current and previous high-level actions, respectively.
The exploration bonus $r_{\text{CAI}}$ is derived from the CAI measure $C^j$ from \eqref{eq:pointwise_cmi}, where we choose the object to be the entity of interest $j$.
% To incentivize influence over the object, we select its position $\boldsymbol{s}_o$ as $S_j$.
Resorting to an approximation $\Tilde{C}_j$, we compute the reward $r_{\text{CAI}}$ at a given state $S=\boldsymbol{s}$ as
\begin{equation}
    \begin{aligned}
        &r_{\text{CAI}} = \Tilde{C}_{j=\text{object}}(s) = \\
        &\frac{1}{K} \sum_{i=1}^K D_{\mathrm{KL}} \left( P_{S_j' \mid S=s, A=a^{(i)}} \bigg\| \frac{1}{K} \sum_{k=1}^K P_{S_j' \mid S=s, A=a^{(k)}} \right),
    \end{aligned}
\label{eq:cai}
\end{equation}
given \( K \) actions \( \{a^{(i)}\}_{i=1}^K \) sampled from the policy. In this work, we find it sufficient to use $K=64$.
This approximation estimates the marginal $P_{S_j' \mid s}$ using Monte Carlo sampling.
We model the transition distribution $P_{S_j' \mid S=s, A=a}$ as a fixed-variance Gaussian (details in~\Cref{subsec:dynamics_learning}), which enables closed-form KL divergence calculation using the Gaussian mixture approximation~\citep{durrieu2012lower}.

The CAI reward encourages the agent to reach states where it exerts greater influence over the object, thereby promoting task-relevant exploration and learning.
The weight $w_2$ for the CAI reward scales with the raw CAI score:
\begin{equation}
    w_{2} = w_{2, b} + \max\left(0, (r_{\text{CAI}} - \alpha_1)/\alpha_2\right),
\label{eq:cai_weight}
\end{equation}
where $\alpha_1$ is the threshold for scaling and $\alpha_2$ controls the scaling rate.
% This adaptive weighting further incentivizes exploration in states with higher causal influence.

Finally, we foster more directed exploration by injecting time-correlated noise into the action sampling process during training~\cite{eberhard2023pink, hollenstein2024colored}. 
Sampling from time-correlated actions reduces the possibility of meaningless back-and-forth behavior that could result from commonly used white-noise samples. 
We select a correlation strength parameterized as $\beta = 0.5$, corresponding to a colored noise between white and pink. 
%For more details on training the high-level policy, we refer the reader to~\Cref{app:additional_training}.

\subsection{Dynamics learning}
\label{subsec:dynamics_learning}
To calculate the exploration bonus $r_{\text{CAI}}$ in \eqref{eq:cai}, we learn the transition model $P_{S_j' \mid S=s, A=a^{(k)}}$ for all entities $j$.  In our current setting, we focus on a single entity—the pushable object—but the framework naturally extends to multiple entities. Concretely, this reduces to learning the object transition model $P_{\boldsymbol{s}'_o \mid \boldsymbol{s}_h, \boldsymbol{a}_h}$.
The object state $\boldsymbol{s}_o$ is defined as the object's position, while the high-level state $\boldsymbol{s}_h = (\boldsymbol{\xi}_r, \boldsymbol{\xi}_o, \boldsymbol{v}_r, \boldsymbol{v}_o)$ includes the robot's pose $\boldsymbol{\xi}_r$ and velocity $\boldsymbol{v}_r$ and the object's pose $\boldsymbol{\xi}_o$ and velocity $\boldsymbol{v}_o$.
The high-level action $\boldsymbol{a}_h$ corresponds to the desired robot velocity, and the next object state $\boldsymbol{s}'_o = \boldsymbol{p}'_o = (x'_o, y'_o)$ denotes its position at the subsequent timestep.
Instead of using a pretrained model, \method leverages data collected from high-level interactions during training to efficiently learn the dynamics.
As described earlier, we model the object's transition probability $P_{\boldsymbol{s}'_o \mid \boldsymbol{s}_h, \boldsymbol{a}_h}$ as a fixed-variance Gaussian distribution $\mathcal{N}(\boldsymbol{s'}_o; f_\theta(\boldsymbol{s}_h,\boldsymbol{a}_h), \sigma^2)$, where $f_\theta$ is a neural network that predicts the mean of the distribution.

To enhance learning efficiency, we incorporate a simple kinematic prior model $f_{\text{kin}}$, which estimates the next object position $\Tilde{\boldsymbol{p}}'_o$ using geometric reasoning based on the relative pose between robot and object and the commanded velocity.
This estimate is computed by projecting the robot's velocity $\boldsymbol{a}_h$ onto the direction from the robot to the object and updating the object's position accordingly:
\begin{equation}
\Tilde{\boldsymbol{p}}'_o = 
\begin{cases}
    \begin{aligned}
        &\boldsymbol{p}_o +  \delta t \cdot(\boldsymbol{a}_{h,xy}\cdot\delta\hat{\boldsymbol{p}})\cdot\delta\hat{\boldsymbol{p}},  \\
        &\quad\quad \text{if } \norm{\boldsymbol{p}_o - \boldsymbol{p}_r}_2 \leq \epsilon_p \text{ and } 
        \boldsymbol{a}_{h,xy}\cdot\delta\hat{\boldsymbol{p}} > 0
    \end{aligned} \\
    \boldsymbol{p}_o ,  \  \text{otherwise }
\end{cases}
\label{eq:kin}
\end{equation}
Here, $\delta\hat{\boldsymbol{p}} = (\boldsymbol{p}_o - \boldsymbol{p}_r)/\norm{\boldsymbol{p}_o - \boldsymbol{p}_r}_2$ is the unit vector pointing from the robot to the object, $\delta t$ is the high-level control step size, and $\epsilon_p$ is a distance threshold.
The conditions ensure that the robot is close enough to and moving toward the object.

We combine the kinematic prior with a learned residual model $f_{\text{res}}$, parameterized by $\theta$, to capture complex physical interactions beyond the kinematics model, including nonlinear effects such as friction, drag, and collisions with obstacles.
The final dynamics model is thus:
\begin{equation}
    f_\theta(\boldsymbol{s}_h, \boldsymbol{a}_h) = f_{\text{kin}}(\boldsymbol{s}_h, \boldsymbol{a}_h) + f_{\text{res}}(\boldsymbol{s}_h, \boldsymbol{a}_h; \theta),
\end{equation}
where $f_{\text{kin}}$ is deterministic and independent of $\theta$. 
We train the residual model by minimizing the mean squared error between the predicted and true object positions:
\begin{equation}
    \mathcal{L}_{\text{res}}(\theta) = \frac{1}{N} \sum_{i=1}^{N} \norm{\Tilde{\boldsymbol{s}}'^{(i)}_o - \boldsymbol{s}'^{(i)}_o }_2^2,
\label{eq:res_loss}
\end{equation}
where $\Tilde{\boldsymbol{s}}'^{(i)}_o = f_{\text{kin}}(\boldsymbol{s}^{(i)}_h, \boldsymbol{a}^{(i)}_h) + f_{\text{res}}(\boldsymbol{s}^{(i)}_h, \boldsymbol{a}^{(i)}_h; \theta)$.

Although the high-level action $\boldsymbol{a}_h$ could, in principle, be sampled from the full support of the desired velocity $\boldsymbol{v}_{r,des}$~\cite{seitzer2021causal}, not all commands are feasible for the low-level controller to execute under the robot's current state.
Therefore, we define $\boldsymbol{a}_h = \Tilde{\boldsymbol{v}}_r$ to be the achievable velocity, and use $\Tilde{\boldsymbol{v}}_r$ for both dynamics learning and CAI computation.
Given training samples $\mathcal{D} = {(\boldsymbol{s}_h^{(i)}, \boldsymbol{a}_h^{(i)}, \boldsymbol{s}'^{(i)}_h)}$, we have access to the achieved robot velocity $\boldsymbol{v}'_r$ at the next timestep.
When computing CAI as an exploration reward, we assume that the robot's velocity can only change within a limited range over one high-level step.
Thus, for any state $\boldsymbol{s}_h$, we sample the action $\boldsymbol{a}_h$ from a bounded range centered on the current velocity:
\begin{equation}
    \boldsymbol{a}_h = \Tilde{\boldsymbol{v}}_{r} \sim \mathcal{U}[\boldsymbol{v}_r \pm \delta \boldsymbol{v}_r]
\end{equation}
where $\delta \boldsymbol{v}_r$ is a fixed range of velocity deviation.
The limits were empirically determined to be $(\delta v_r^x, \delta v_r^y, \delta \omega_r^z) = (0.3, 0.3, 0.4)$. 
% Detailed values for the velocity limits $(\delta v_r^x, \delta v_r^y, \delta \omega_r^z)$ are provided in~\Cref{app:hyperparams}.

\begin{figure*}[t!]
    \centering
    \begin{minipage}{0.32\textwidth}
        \centering
        \includegraphics[width=\linewidth]{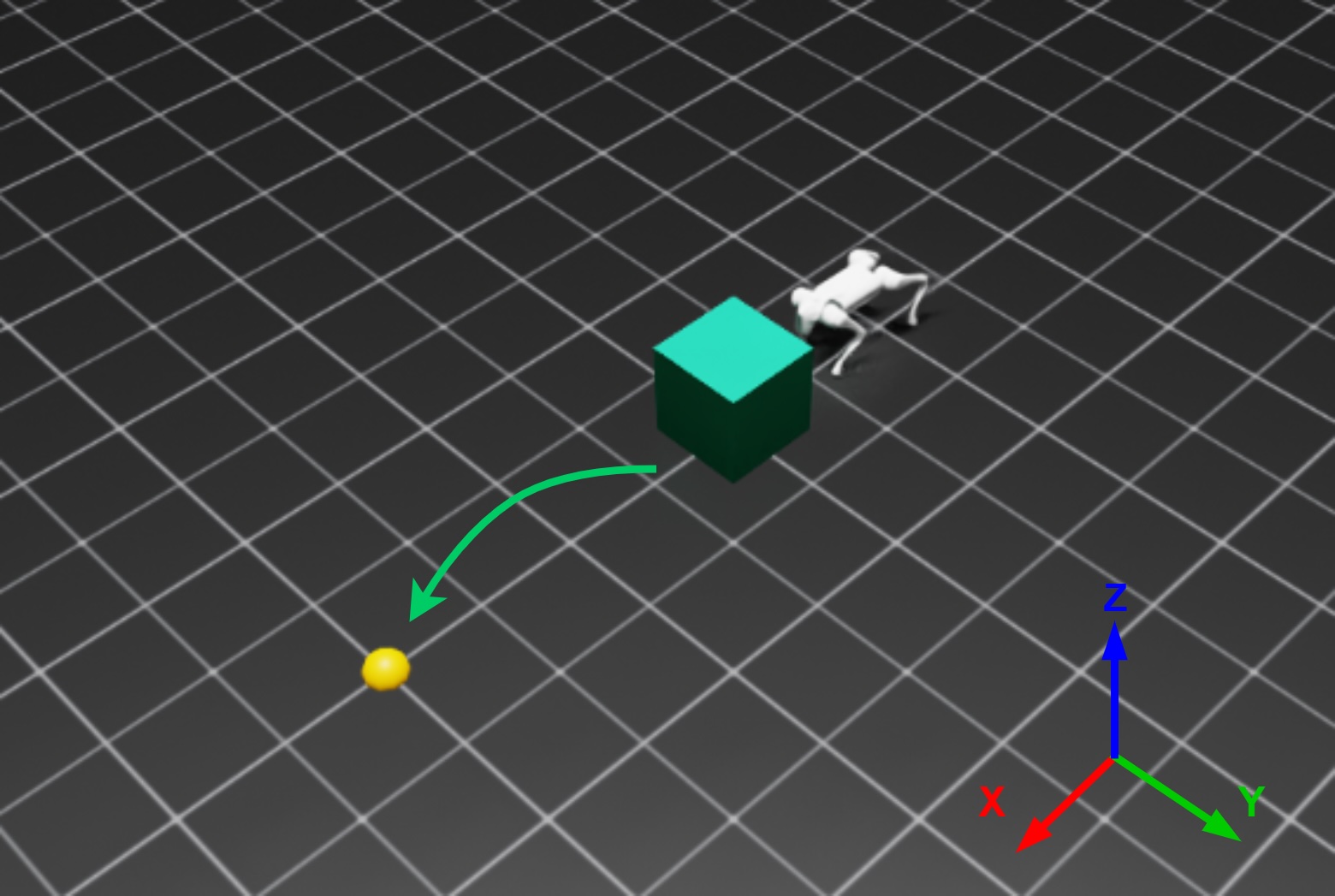}
        % \caption*{a}
    \end{minipage}%
    \hspace{0.01\textwidth}
    \begin{minipage}{0.32\textwidth}
        \centering
        \includegraphics[width=\linewidth]{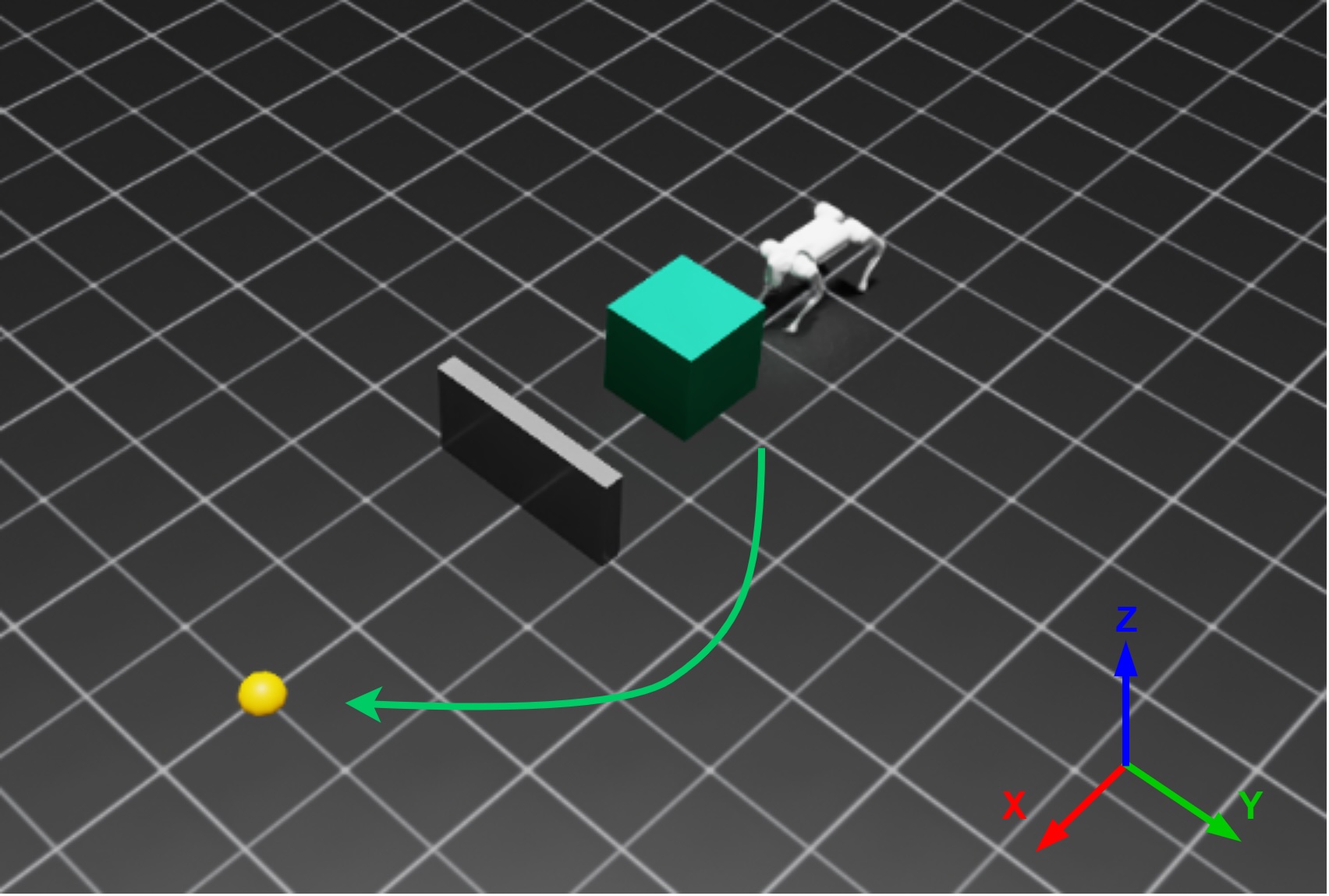}
        % \caption*{b}
    \end{minipage}%
    \hspace{0.01\textwidth}
    \begin{minipage}{0.32\textwidth}
        \centering
        \includegraphics[width=\linewidth]{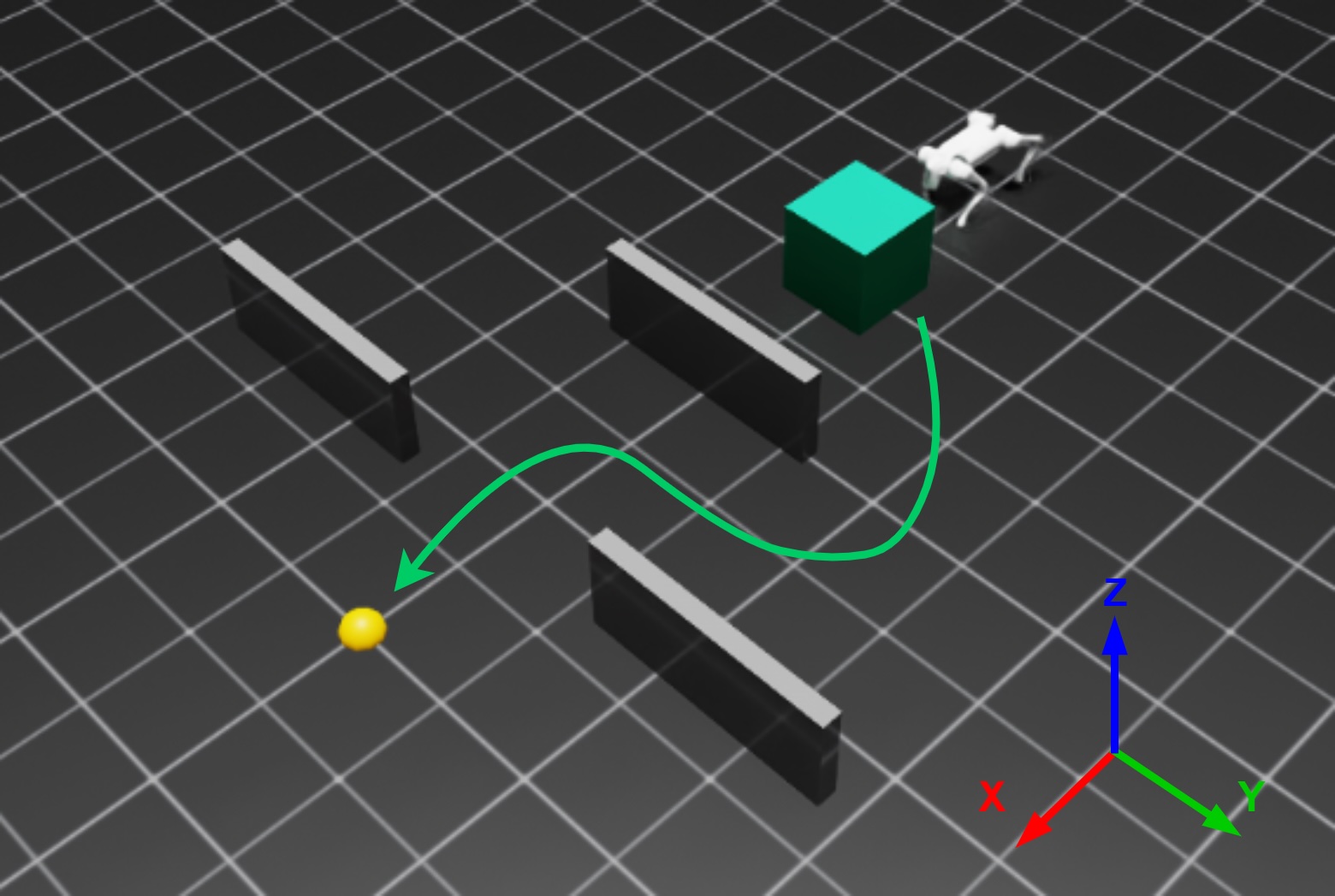}
        % \caption*{c}
    \end{minipage}
    % Overall caption for the entire figure
    \caption{The \code{Single Object} (\textit{left}), \code{Single Wall} (\textit{middle}), and \code{Multi-Wall} (\textit{right}) tasks. The yellow sphere denotes the object's target position.}
    
    % Label for referencing the figure
    \label{fig:tasks}
    \vspace{-0.6cm}
\end{figure*}

\section{EXPERIMENTS}

We trained both high- and low-level policies using Isaac Sim~\cite{mittal2023orbit}.
%Training hyperparameters and additional details are provided in~\Cref{app:additional_training}.
The low-level policy operates with a control interval of 0.02 seconds, while the high-level policy has a control interval of 0.2 seconds. 
At each iteration of the high-level control loop, we advance the environment by 0.2 seconds, compute rewards and terminations, collect transitions for both the high-level and dynamics policies, and augment the rewards buffer with the CAI exploration bonus.

We evaluated \method on three pushing tasks of increasing difficulty, illustrated in~\Cref{fig:tasks}.
In the \code{Single-object} task, the robot pushes a single cuboid object to a fixed target position. The \code{Single-wall} task introduces a fixed wall that blocks the direct path to the target, requiring the robot to maneuver the object around the obstacle. The \code{Multi-wall} task further increases complexity with multiple fixed walls through which the robot must navigate the object. 
The wall obstacles in the single-wall and multi-wall tasks introduce regions in the state space where the object's dynamics become more complex, thereby reducing the robot's ability to exert consistent influence over the object.
We trained 1500 iterations for the single object and single wall tasks, and 2000 iterations for the multi-wall task, where each iteration consists of 10 high-level control steps. 
All tasks have an episode duration of 20 seconds and use a fixed target position.
%A full description of scene configurations is provided in~\Cref{app:scene_configs}.
% ADD SOME MORE TRAINING DETAILS HERE (iterations trained) AND TO THE FRAMEWORK IMAGE

\begin{table}[t]
    \vspace{0.2cm}
    \centering
    \setlength{\tabcolsep}{4pt} % Default value: 6pt
    \renewcommand{\arraystretch}{1.2} % Default value: 1
    \begin{tabular}{|c|c|}
         \hline
         \multicolumn{2}{|c|}{\textbf{Reward functions}} \\ \hline
         CAI &$r = w_1r_{\text{task}} + w_2r_{\text{CAI}} + w_3r_{\text{reg}} $ \\
         \hline
         RND & $r = w_1r_{\text{task}} + w_4r_{\text{RND}} + w_3r_{\text{reg}} $ \\
         \hline
         heuristics & $r = w_1r_{\text{task}} + w_5r_{\text{heu}} + w_3r_{\text{reg}} $ \\
         \hline
         $r_{\text{CAI}}$ weight & \Cref{eq:cai_weight} \\
         \hline
    \end{tabular}
    
    \vspace{2pt}
    \begin{tabular}{|c|c|c|c|c|c|c|c|}
        \hline
         \multicolumn{8}{|c|}{\textbf{Reward coefficients}} \\ \hline
         \textbf{Task} & \textbf{$w_1$} & \textbf{$w_3$} & \textbf{$w_4$} & \textbf{$w_5$} & \textbf{$w_{2, b}$} & \textbf{$\alpha_1$} & \textbf{$\alpha_2$}\\ \hline
         Single object & 15 & -5e-3 & 10 & 0.01 & 40 & 12e-5 & 1.5e-6 \\ \hline
         Single wall & 40 & -5e-3 & 10 & 0.01 & 40 & 12e-5 & 1.5e-6 \\ \hline
         Multi-wall & 40 & -5e-3 & 10 & 0.01 & 40 & 12e-5 & 1.5e-6 \\ \hline
    \end{tabular}
    \vspace{0.1cm}
    \caption{Simulation training reward parameters. All CAI methods (\code{CAI-kinematic, CAI-learned}, \method) use the CAI reward function, all RND methods (\code{RND-full, RND-object}) use the RND reward function, while only the \code{Heuristics} method uses the heuristics reward function.}
    \label{tab:rewards}
    \vspace{-0.6cm}
\end{table}

We compared our approach against several baselines.
The \code{Heuristics} baseline trains the high-level policy using a distance-based heuristic reward, \mbox{$r_{\text{heu}} = \exp\left(-\norm{\boldsymbol{p}_r - \boldsymbol{p}_o}_2\right)$}, which encourages the robot to minimize its distance to the object and mirrors the primary exploration reward used in prior work~\citep{jeon2023learning}.
We further evaluated \method against intrinsic motivation algorithms, specifically Random Network Distillation
(RND)~\citep{burda2018exploration}, implementing two variants: \code{RND-full}, which follows the original formulation and computes prediction error over the full state, and \code{RND-object}, which restricts prediction error to the object state only.
For ablation studies, we introduced
\code{CAI-kinematic}, where CAI is computed using only the kinematic prior (\Cref{eq:kin}), and \code{CAI-learned}, where CAI relies on a fully learned dynamics model (no prior).
The reward functions and corresponding parameters for all tasks and training methods are summarized in \Cref{tab:rewards}. The RND reward term is given as $r_{RND} = \lVert f(s) - \hat{f}(s) \rVert_2$, where $s$ is the state to be encoded, and $f(s)$ and $\hat{f}(s)$ denote the target and predictor networks, respectively.

\begin{figure*}[ht!]
    \centering
        %\centering2
         \begin{minipage}{0.32\textwidth}
            \centering
            \includegraphics[width=\linewidth]{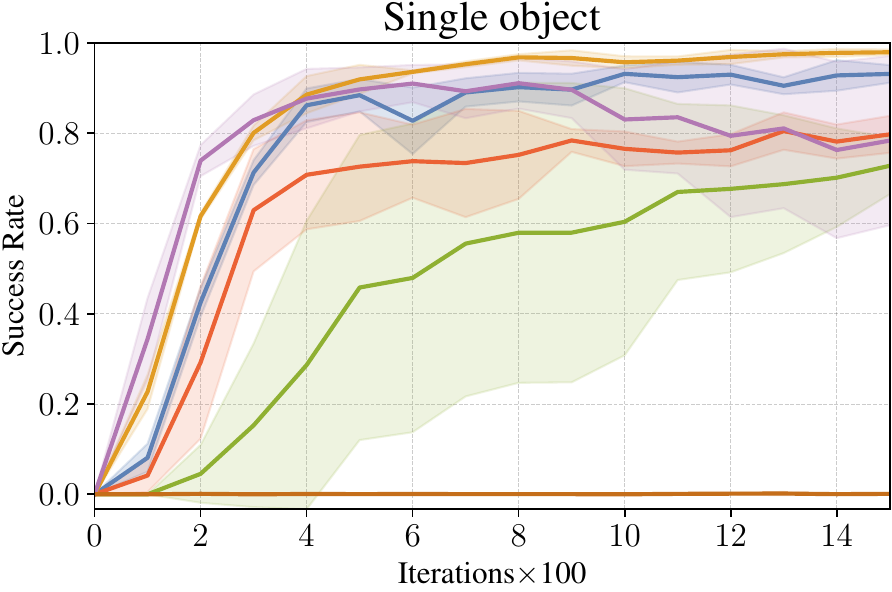}
        \end{minipage}%
        \hspace{0.01\textwidth}  % Space between images
        \begin{minipage}{0.32\textwidth}
            \centering
            \includegraphics[width=\linewidth]{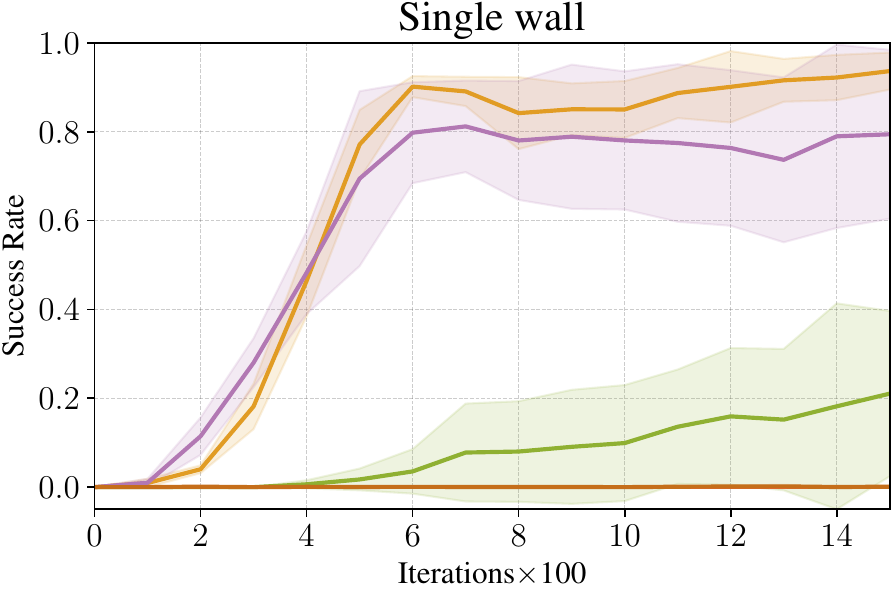}
        \end{minipage}%
        \hspace{0.01\textwidth}  % Space between images
        \begin{minipage}{0.32\textwidth}
            \centering
            \includegraphics[width=\linewidth]{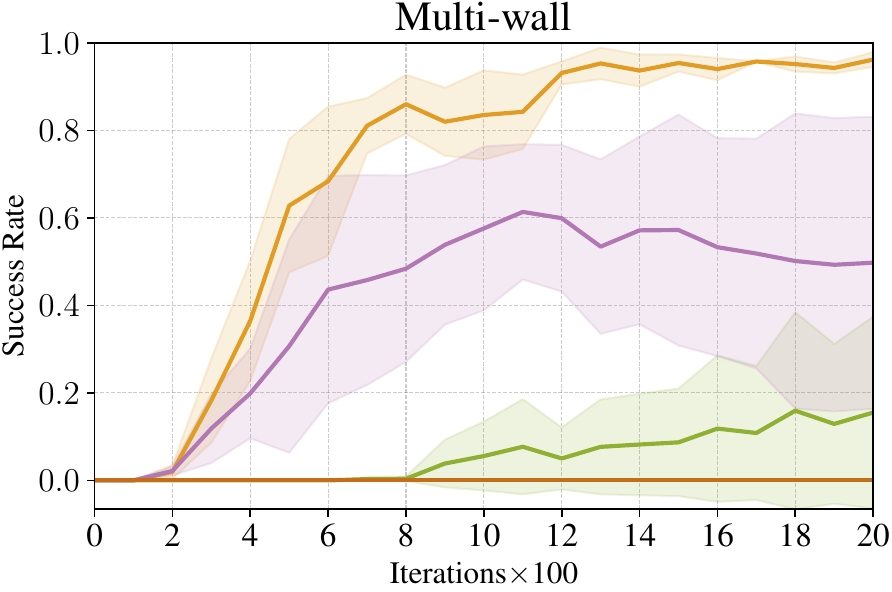}
        \end{minipage}

            % Legend
        \vspace{.5em} % Space between images and legend
        
        \begin{tikzpicture}
            % Legend entries
            \draw[draw=none, fill=heuristics]
                (0,0.1) rectangle (0.5,0.2); % rectangle
            \node[right] at (0.5,0.15) {\footnotesize{Heuristics}}; % text
            \draw[draw=none, fill=caiprior]
                (3.0,0.1) rectangle (3.5,0.2); % rectangle
            \node[right] at (3.5,0.15) {\footnotesize{CAI-kinematic}}; % text
            \draw[draw=none, fill=cailearn]
                (6.0,0.1) rectangle (6.5,0.2); % rectangle
            \node[right] at (6.5,0.15) {\footnotesize{CAI-learned}}; % text
            \draw[draw=none, fill=ourbrown]
                (0.0,-0.1) rectangle (0.5,-0.2); % rectangle
            \node[right] at (0.5,-0.15) {\footnotesize{RND-full}}; % text
            \draw[draw=none, fill=ourviolet]
                (3.0,-0.1) rectangle (3.5,-0.2); % rectangle
            \node[right] at (3.5,-0.15) {\footnotesize{RND-object}}; % text
            \draw[draw=none, fill=caiman]
                (6.0,-0.1) rectangle (6.5,-0.2); % rectangle
            \node[right] at (6.5,-0.15) {\footnotesize{\method}}; % text
        \end{tikzpicture}
    
    % Overall figure caption
    \caption{Policy success rate evaluated at every 100 training iterations for all methods and tasks. Results are evaluated across 800 episodes and averaged over 3 seeds, shaded area represents standard deviation. 
}
    \label{fig:generalsuccess}
    \vspace{-0.6cm}
\end{figure*}

\subsection{Simulation results}
We evaluated the learning performance of each method by measuring the success rate for each task.
A task is successfully completed if the distance between the object and the target is below a threshold $\epsilon_s$ at the end of an episode.
The results are shown in~\Cref{fig:generalsuccess}.

With only sparse task rewards, task-relevant skill acquisition heavily relies on effective exploration.
In the \code{Single-object} task, all methods---except \code{RND-full}---achieve some degree of success.
\method, \code{RND-object}, and \code{Heuristics} perform the best, while \code{CAI-kinematic} and \code{CAI-learned} underperform.
The low performance of \code{CAI-learned} is likely due to the high data and time demands required to learn an accurate dynamics model from scratch.
In contrast, \code{CAI-kinematic} performs comparably with the top methods, suggesting that the naive kinematic prior is sufficient for simpler tasks with straightforward object interactions.
In the more complex \code{Single-wall} and \code{Multi-wall} tasks, only \method and \code{RND-object} demonstrate notable success, with \code{RND-object} showing lower asymptotic performance than \method in both scenarios.
This highlights \method's unique ability to guide task-relevant exploration even in the presence of obstacles, validating its effectiveness in cluttered environments and independence from dense extrinsic rewards.
The failure of \code{CAI-kinematic} and \code{CAI-learned} in these tasks emphasizes the importance of combining a kinematic prior with a learned residual model to efficiently capture the complex object dynamics needed for effective exploration.

\code{RND-full} results in agents randomly exploring the state space without yielding meaningful behavior in all tasks. This finding reinforces the need for object-centric exploration bonuses when the skill to be learned involves physical interaction with objects.
The performance gap between \code{RND-object} and \method can be partially attributed to the phenomenon of \emph{detachment}~\cite{ecoffet2019go}, where the agent drifts away from reward-depleted areas and continuously pursues new regions with higher intrinsic reward, potentially losing avenues of task-relevant exploration.

\begin{figure}[t!]
    \vspace{0.2cm}
    \centering
    \begin{minipage}{0.49\textwidth}
        \centering
        \includegraphics[width=\linewidth]{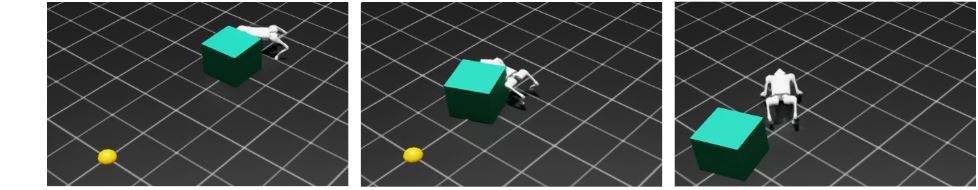}
    \end{minipage}
    \begin{minipage}{0.49\textwidth}
        \centering
        \includegraphics[width=\linewidth]{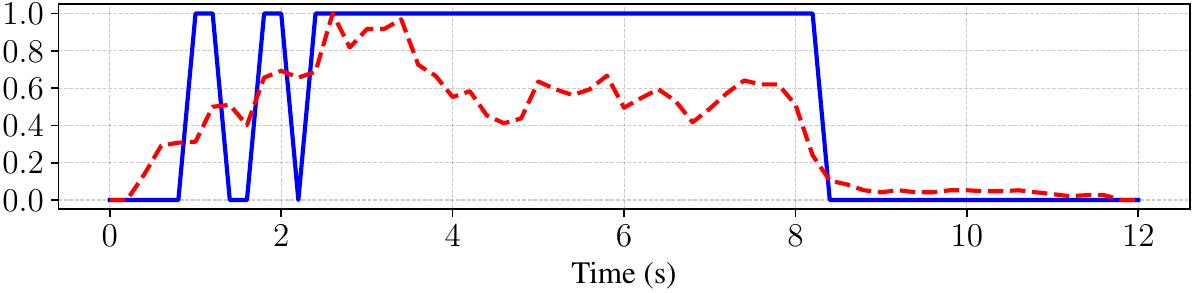}
    \end{minipage}
    
    \begin{tikzpicture}
        \draw[red, thick, line width = 0.9pt, dash pattern = on 3pt off 1.5pt] (1.5,0.15) -- (2.0,0.15);
        \node[right] at (2.0,0.15) {\footnotesize{CAI reward}};
    
        \draw[draw=none, fill=blue] (4.5,0.1) rectangle (5.0,0.2);
        \node[right] at (5.0,0.15) {\footnotesize{contact}};
    \end{tikzpicture}
    \caption{Scaled CAI reward overlaid with a binary indicator of robot–object contact (1-contact, 0-no contact) over one episode of the \code{Single-object} task. Representative environment frames are shown above.}
    \label{fig:cai_vs_contact}
    \vspace{-0.6cm}
\end{figure}

To validate CAI as an effective exploration signal, \Cref{fig:cai_vs_contact} explicitly plots the raw CAI reward alongside robot-object interactions in the \code{Single-object} task. Elevated CAI values during contact states empirically confirm the role of CAI in guiding exploration.

\subsection{Loco-manipulation of irregular objects}

\begin{figure}[t!]
% \vspace{-0.5cm}
    \centering
    \begin{minipage}{0.18\textwidth}
        \centering
        \includegraphics[width=\linewidth]{
            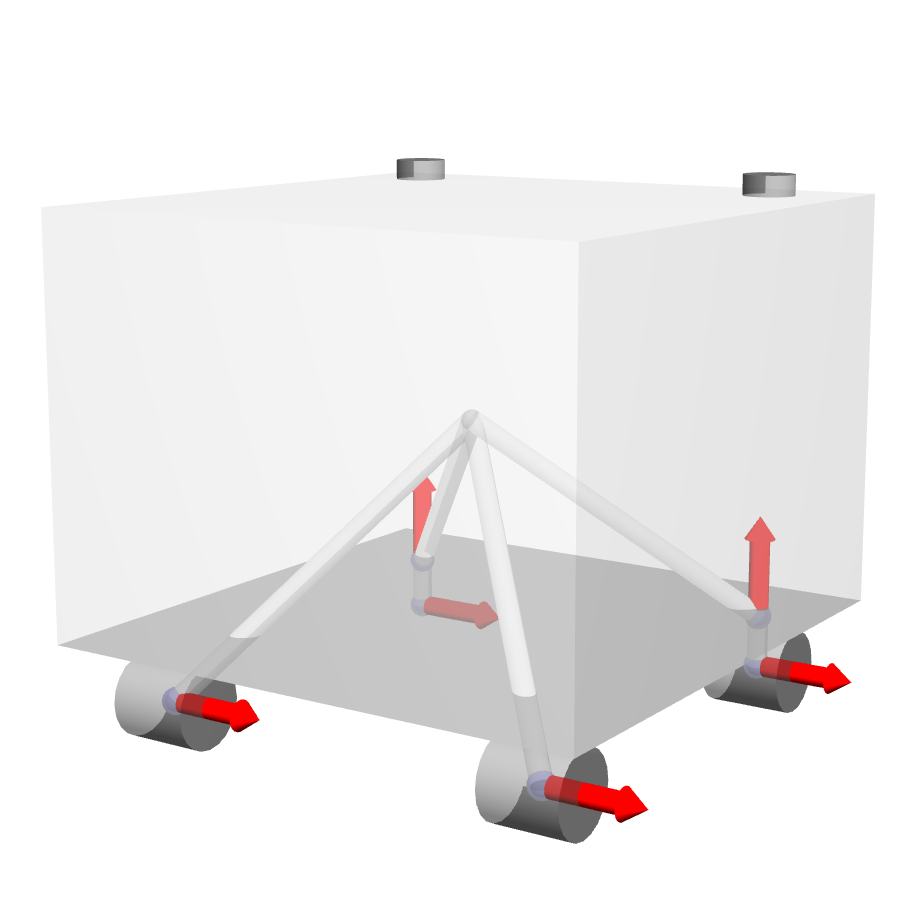
        }
    \end{minipage}%
    \hspace{0.001\textwidth}
    \begin{minipage}{0.25\textwidth}
        \centering
        \includegraphics[width=\linewidth]{
            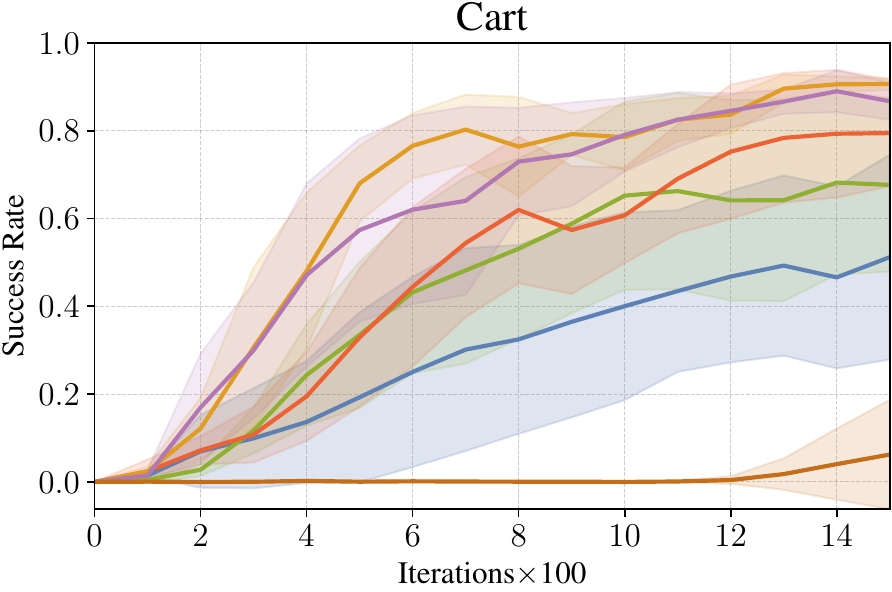
        }
    \end{minipage}%
    % \vspace{1em}

    \begin{tikzpicture}
        % Legend entries
        \draw[draw=none, fill=heuristics]
            (0,0.1) rectangle (0.5,0.2); % rectangle
        \node[right] at (0.5,0.15) {\footnotesize{Heuristics}}; % text
        \draw[draw=none, fill=caiprior]
            (3.0,0.1) rectangle (3.5,0.2); % rectangle
        \node[right] at (3.5,0.15) {\footnotesize{CAI-kinematic}}; % text
        \draw[draw=none, fill=cailearn]
            (6.0,0.1) rectangle (6.5,0.2); % rectangle
        \node[right] at (6.5,0.15) {\footnotesize{CAI-learned}}; % text
        \draw[draw=none, fill=ourbrown]
            (0.0,-0.1) rectangle (0.5,-0.2); % rectangle
        \node[right] at (0.5,-0.15) {\footnotesize{RND-full}}; % text
        \draw[draw=none, fill=ourviolet]
            (3.0,-0.1) rectangle (3.5,-0.2); % rectangle
        \node[right] at (3.5,-0.15) {\footnotesize{RND-object}}; % text
        \draw[draw=none, fill=caiman]
            (6.0,-0.1) rectangle (6.5,-0.2); % rectangle
        \node[right] at (6.5,-0.15) {\footnotesize{\method}}; % text
    \end{tikzpicture}
    %\vspace{-0.1cm}
    \caption{\textit{Left}: A simplified cart object. \textit{Right}: Single
    cart pushing results.}
    % mark cart sparse no wall
    \label{fig:cart}
    \vspace{-0.6cm}
\end{figure}

To demonstrate \method's effectiveness in handling complex objects with asymmetric dynamics, we conducted an additional experiment using a simplified cart
with two normal and two caster wheels, as illustrated in~\Cref{fig:cart}. The scene configuration is identical to that of the \code{Single-object} task, with the cuboid being replaced by the cart. 
Our results show that \method is more sample efficient and
achieves a higher asymptotic performance (approx. 90\% success rate) than most baselines and is on par with the best competitor. In particular, this experiment did not require any changes to the original framework or to the kinematics prior, which was derived assuming an object with symmetric motion. These results highlight the ability of the learned model to accurately capture complex dynamics and confirm \method's capability in learning with irregular objects. This environment has no obstacles, but we hypothesize that adding walls would further emphasize \method's advantages over other baselines.
\subsection{Leveraging pretrained dynamics for new tasks}
Since our framework learns the environment dynamics, we propose that the dynamics residual obtained from a previous training can be reused across different tasks---as long as the underlying dynamics remain consistent.
We hypothesize that leveraging a pretrained model improves the accuracy of CAI estimates in the early stages of training, thus enhancing the exploration signal and significantly boosting sample efficiency.

\begin{figure}[t]
    % \vspace{-0.5cm}
        \centering
        \begin{minipage}{0.18\textwidth}
            \centering
            \includegraphics[width=\linewidth]{
                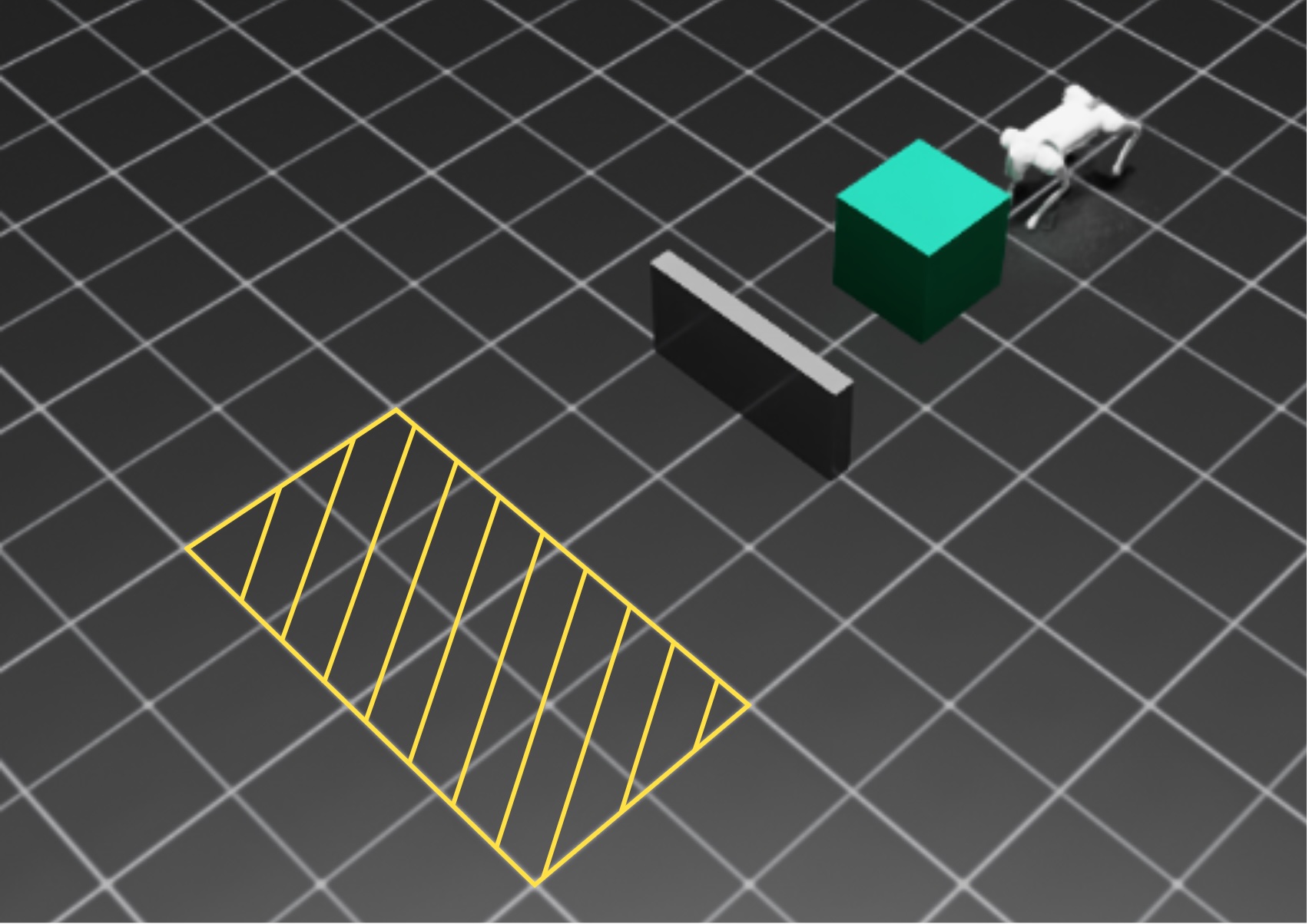
            }
        \end{minipage}%
        \hspace{0.001\textwidth}
        \begin{minipage}{0.25\textwidth}
            \centering
            \includegraphics[width=\linewidth]{
                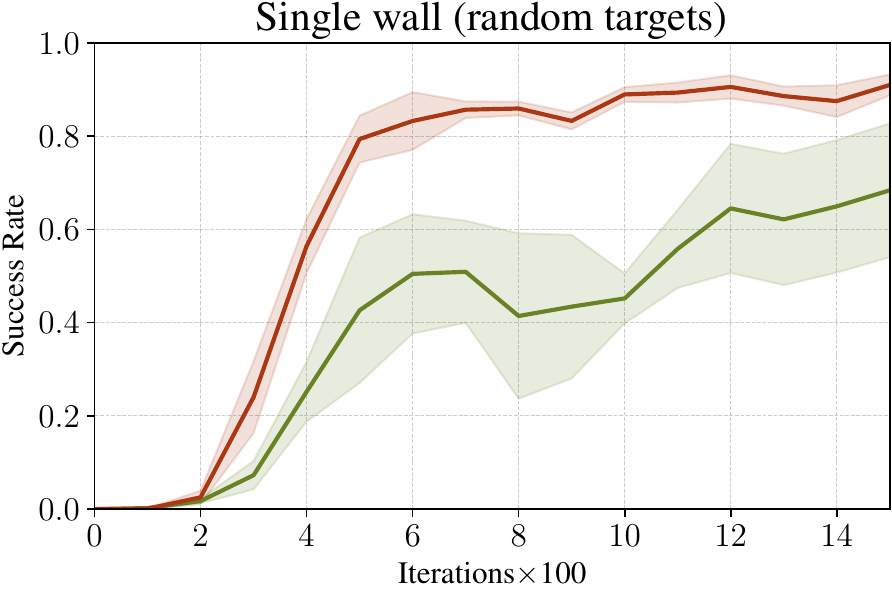
            }
        \end{minipage}%\
        
        \begin{tikzpicture}
            % Legend entries
            \draw[draw=none, fill=ourdarkred]
                (1.5,0.1) rectangle (2.0,0.2); % rectangle
            \node[right] at (2.0,0.15) {\footnotesize{with pretrained}}; % text
            \draw[draw=none, fill=ourdarkgreen]
                (4.5,0.1) rectangle (5.0,0.2); % rectangle
            \node[right] at (5.0,0.15) {\footnotesize{w/o pretrained}}; % text
        \end{tikzpicture}
        \caption{\textit{Left}: Single wall task with target positions randomly sampled within an area in front of the wall. \textit{Right}: Learning the single wall random target task using models without pretraining (learned from scratch) and models pretrained from the fixed target task.}
        \label{fig:transfer_dynamics}
        \vspace{-0.5cm}
\end{figure}

To test this hypothesis, we consider a generalization of the \code{Single-wall} task in which the target position is randomized instead of fixed, sampled uniformly within a predefined area as illustrated in~\Cref{fig:transfer_dynamics}.
%Detailed scene configurations are provided in~\Cref{app:scene_configs}.
We compare the original \method, which learns the dynamics residual from scratch, to one that reuses a pretrained residual model. For the latter, we initialize the residual with one from the original \code{Single-wall} task and continuously fine-tune it during training with data collected from the new generalized task.

As shown in~\Cref{fig:transfer_dynamics}, the reuse of learned dynamics significantly accelerates learning with sparse rewards.
The pretrained model yields more informative CAI estimates, guiding the robot toward meaningful interaction behaviors early in training.
This leads to substantial gains in sample efficiency, supporting our hypothesis and demonstrating the benefits of transferring learned dynamics across related tasks.

\subsection{Hardware deployment}
\begin{figure}[t]
    \centering
    \begin{subfigure}[b]{\linewidth}
    \includegraphics[width=\linewidth]{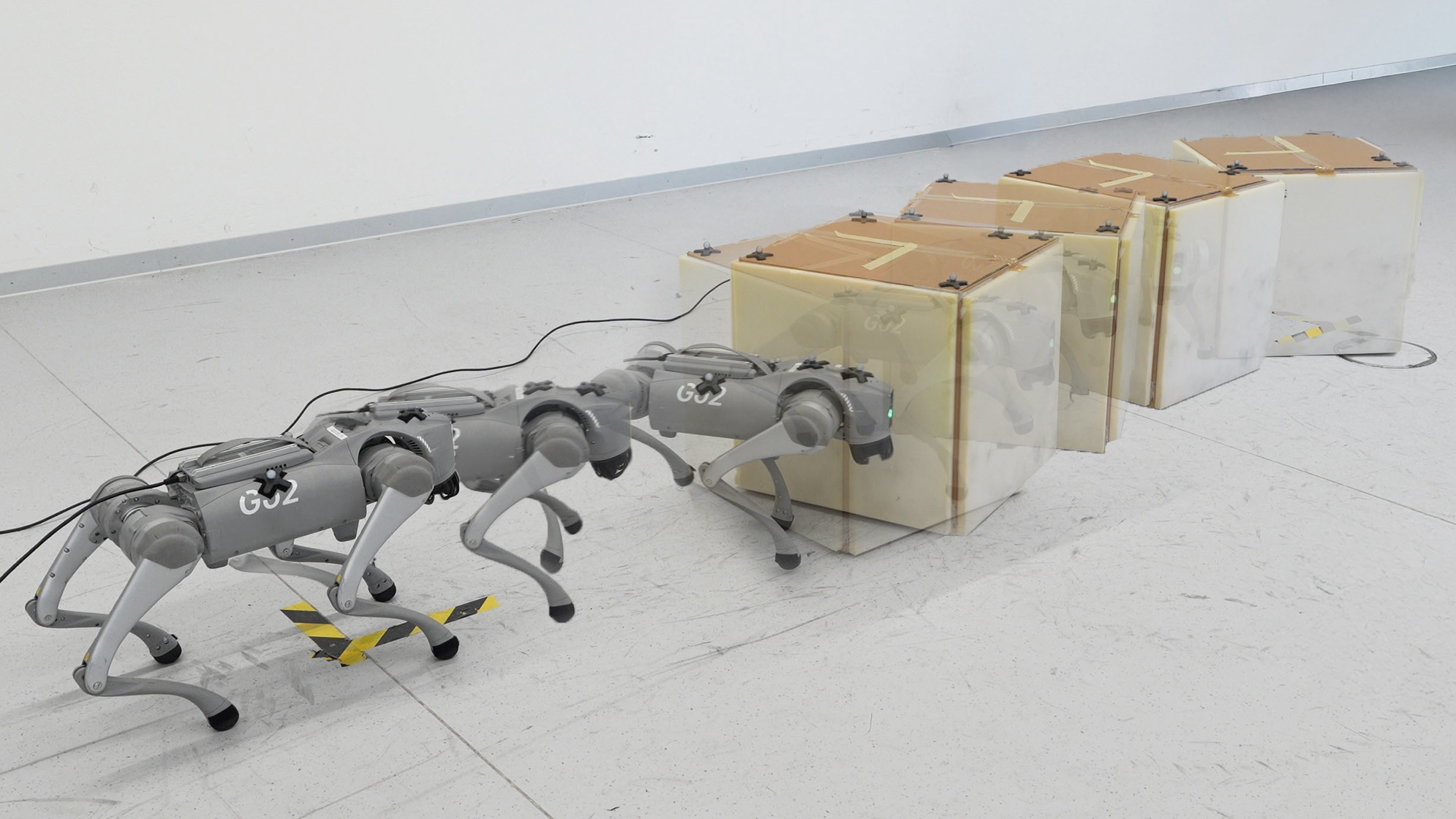}
    \end{subfigure}
    \par\medskip
    \begin{subfigure}[b]{\linewidth}
    \includegraphics[width=\linewidth]{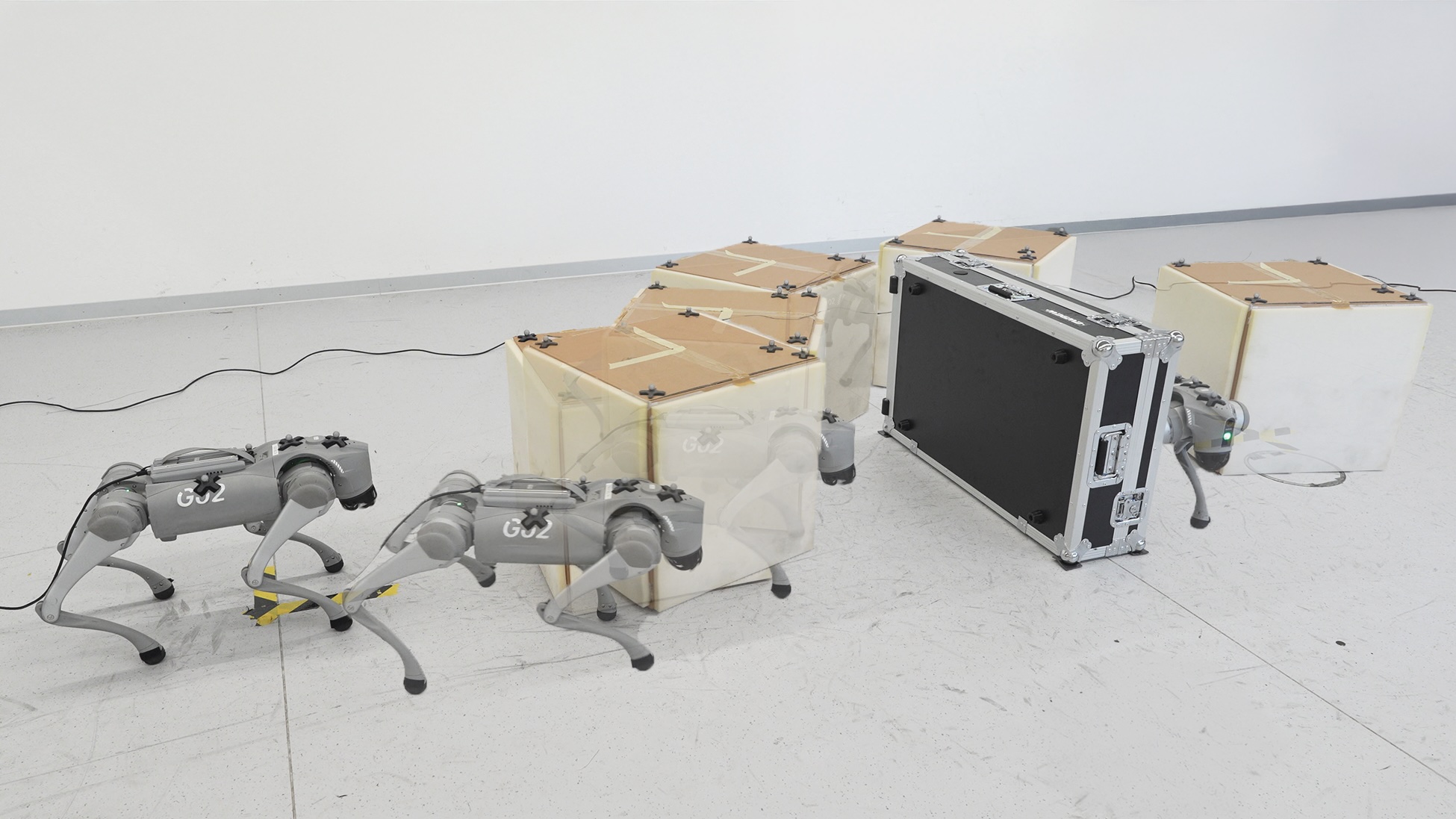}
    \end{subfigure}

    \caption{Snapshots from hardware deployment for the single object and single wall tasks. The mass of the box is 5.5 kilograms with a dimension of $(0.55, 0.55, 0.5)$ meters. } 
    \label{fig:Hardware}
    \vspace{-0.5cm}
\end{figure}

We validated our trained policies on real-world quadruped pushing tasks with the Unitree Go2, as shown in~\Cref{fig:Hardware}.
To bridge the sim-to-real gap, we applied domain randomization to the object's mass and friction during high-level policy training.
%Additional details are provided in~\Cref{app:dr_hw}.
To obtain the observations for the high-level policies, all entities in the environment were tracked using an external motion capture system.
Our trained policies were directly deployed to the robot and successfully executed pushing tasks without requiring additional fine-tuning.
For further details and visualizations of our trained policies in simulation and on hardware, please refer to the supplementary video.

\section{CONCLUSION}
% A conclusion section is not required. Although a conclusion may review the main points of the paper, do not replicate the abstract as the conclusion. A conclusion might elaborate on the importance of the work or suggest applications and extensions. 
We presented \method, a general framework for training whole-body pushing skills in legged robots.
\method adopts a hierarchical control strategy that decouples locomotion from high-level planning and introduces an intrinsic CAI-based exploration bonus, encouraging control over relevant entities without hand-crafted reward shaping.
% By introducing an intrinsic exploration bonus based on CAI, our method encourages the robot to gain control over relevant entities in its environment under sparse task signals---eliminating the need for tedious, hand-crafted reward shaping.
The CAI computation is bootstrapped with a simple yet effective kinematic prior that is refined by a learned residual dynamics model.
% Through extensive simulation experiments in a suite of quadruped pushing tasks, we established that \method consistently outperforms competitive baselines in terms of sample efficiency, particularly in scenarios with obstacles. We also illustrated \method's potential for learning with irregular objects, highlighting the effectiveness of our approach in capturing complex dynamics.
Across diverse quadruped pushing tasks, \method outperforms competitive baselines in sample efficiency, especially in obstacle-rich scenarios, and moreover generalizes to irregular objects.
% In addition, we showed how the learned dynamics residual can be reused to accelerate training for new tasks, thus underscoring the transferability and modularity of our framework.
We also showed that the learned dynamics residual accelerates training in new tasks, highlighting the transferability of our approach. 
Finally, hardware experiments demonstrate the seamless deployment of policies trained with \method on a real quadruped.
Overall, \method provides a robust and scalable solution for physically grounded manipulation behaviors in legged robots, paving the way for more autonomous and adaptable robotic systems.

\section{ACKNOWLEDGMENTS}
The authors thank the Max Planck ETH Center for Learning Systems for supporting Núria Armengol, and thank Dongho Kang, Zijun Hui, and Taerim Yoon for developing the ROS framework used in hardware deployment.


%%%%%%%%%%%%%%%%%%%%%%%%%%%%%%%%%%%%%%%%%%%%%%%%%%%%%%%%%%%%%%%%%%%%%%%%%%%%%%%%

%%%%%%%%%%%%%%%%%%%%%%%%%%%%%%%%%%%%%%%%%%%%%%%%%%%%%%%%%%%%%%%%%%%%%%%%%%%%%%%%



% \section*{ACKNOWLEDGMENT}

% The preferred spelling of the word ÒacknowledgmentÓ in America is without an ÒeÓ after the ÒgÓ. Avoid the stilted expression, ÒOne of us (R. B. G.) thanks . . .Ó  Instead, try ÒR. B. G. thanksÓ. Put sponsor acknowledgments in the unnumbered footnote on the first page.

%%%%%%%%%%%%%%%%%%%%%%%%%%%%%%%%%%%%%%%%%%%%%%%%%%%%%%%%%%%%%%%%%%%%%%%%%%%%%%%%

% References are important to the reader; therefore, each citation must be complete and correct. If at all possible, references should be commonly available publications.

{\footnotesize
\bibliographystyle{IEEEtran}
\bibliography{ICRA2026/IEEEabrv, ICRA2026/references}
}

\end{document}